\documentclass[10pt,twocolumn,letterpaper]{article}

\usepackage{cvpr}              

\definecolor{cvprblue}{rgb}{0.21,0.49,0.74}
\usepackage[pagebackref,breaklinks,colorlinks,allcolors=cvprblue]{hyperref}

\def\paperID{5639} 
\def\confName{CVPR}
\def\confYear{2025}

\usepackage{hyperref}
\hypersetup{
colorlinks = true, 
urlcolor = magenta, 
linkcolor = red, 
citecolor = blue}
\usepackage{multirow}
\usepackage{enumitem}
\usepackage{romannum}
\usepackage{booktabs}
\usepackage{amsmath}
\usepackage{cuted}
\usepackage{capt-of}

\usepackage{array}
\newcolumntype{P}[1]{>{\centering\arraybackslash}p{#1}}

\usepackage{xcolor,colortbl}
\definecolor{Gray}{gray}{0.9}

\title{Transforming Static Images Using Generative Models\\for Video Salient Object Detection}
\author{Suhwan Cho\quad Minhyeok Lee\quad Jungho Lee\quad Sangyoun Lee\vspace{0.5cm}\\
Yonsei University}

\begin{document}
\maketitle

\begin{abstract}
In many video processing tasks, leveraging large-scale image datasets is a common strategy, as image data is more abundant and facilitates comprehensive knowledge transfer. A typical approach for simulating video from static images involves applying spatial transformations, such as affine transformations and spline warping, to create sequences that mimic temporal progression. However, in tasks like video salient object detection, where both appearance and motion cues are critical, these basic image-to-video techniques fail to produce realistic optical flows that capture the independent motion properties of each object. In this study, we show that image-to-video diffusion models can generate realistic transformations of static images while understanding the contextual relationships between image components. This ability allows the model to generate plausible optical flows, preserving semantic integrity while reflecting the independent motion of scene elements. By augmenting individual images in this way, we create large-scale image-flow pairs that significantly enhance model training. Our approach achieves state-of-the-art performance across all public benchmark datasets, outperforming existing approaches. Code and models are available at 
\url{https://github.com/suhwan-cho/RealFlow}.
\end{abstract}

\begin{figure}[t]
\centering
\includegraphics[width=0.85\linewidth]{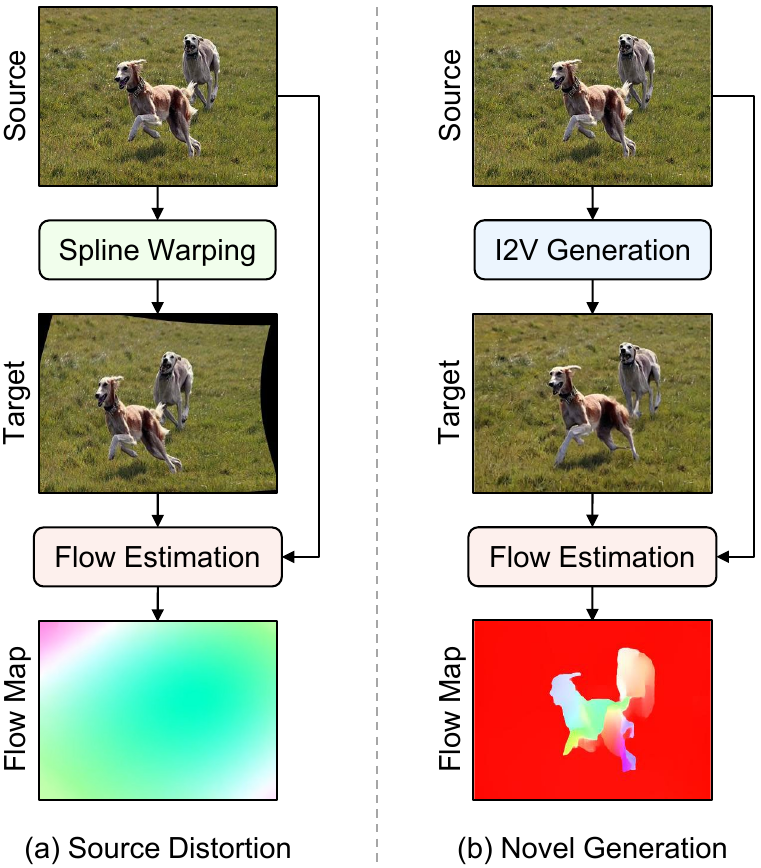}
\caption{Visualization of the optical flow generation process from a static image using two different target image generation methods.}
\label{figure1}
\end{figure}

\section{Introduction}
In deep learning-based image and video processing tasks, the quantity and quality of training data are crucial factors influencing model performance. However, video data introduces additional complexities due to its temporal dimension, making annotation far more labor-intensive and costly compared to image data. To address this challenge, researchers often turn to large-scale, annotated image datasets, which are more accessible and diverse, to support model learning for video tasks. A common strategy for adapting image data to video tasks involves transforming static images and their corresponding annotations (e.g., segmentation masks) to simulate sequences that mimic the temporal progression of video content. Spatial transformations, such as affine transformations and spline warping, are applied to static images to introduce motion cues, thereby generating image sequences that resemble actual video data. This approach has proven effective in video processing, allowing models to learn temporal patterns and structural variations from diverse image data, which helps mitigate the limited availability of video data.

However, this approach falls short in tasks where motion cues are critical. For instance, video salient object detection (VSOD) requires both appearance and motion cues to accurately identify and segment prominent objects in video sequences. VSOD models typically adopt a two-stream approach, incorporating both RGB images and optical flow maps to capture appearance and motion properties simultaneously. The flow information provides context for the dynamics of each object, allowing models to distinguish between independent object movement and background motion. Unfortunately, existing image-to-video simulation methods, which rely on simple spatial transformations, fail to generate realistic optical flows, as shown in Figure~\ref{figure1}~(a). These transformations neglect the contextual relationships between objects in the scene, resulting in flows that fail to accurately reflect the independent motion properties of individual elements. As a result, these methods fall short in tasks like VSOD, where both appearance and motion cues are essential for accurate segmentation.

To overcome this limitation, we propose a novel solution using image-to-video diffusion models to generate realistic transformations of static images. Diffusion models possess the ability to generate novel images while fully understanding the contextual relationships within them. This enables the generation of plausible optical flows by transforming source images while preserving the semantic integrity of the scene and reflecting the independent motion of each object, as depicted in Figure~\ref{figure1}~(b). Unlike traditional spatial transformations, which distort images without accounting for object relationships, our approach maintains these relationships to generate high-quality image-flow pairs. These pairs significantly enhance model training for motion-guided tasks such as VSOD. By augmenting static images with diffusion-based transformations, we can simulate realistic video sequences and generate the corresponding optical flows necessary for effective learning in two-stream models.

We evaluate our method on several public VSOD benchmarks. Unlike previous approaches that rely on complex network architectures, we adopt a simple encoder-decoder framework to demonstrate the effectiveness of our training strategy. Experimental results show that augmenting training data with large-scale image-flow pairs significantly improves model performance. Specifically, we achieve new state-of-the-art $\mathcal{S}$ measure~\cite{smeasure} scores of 94.5\%, 92.6\%, 80.3\%, and 96.2\% on the DAVIS 2016~\cite{DAVIS} validation set, FBMS~\cite{FBMS} test set, DAVSOD~\cite{SSAV} test set, and ViSal~\cite{ViSal} dataset, respectively.

Our main contributions can be summarized as follows:
\begin{itemize}[leftmargin=0.2in]
\item We identify the limitations of current image-to-video simulation methods for motion-guided tasks, emphasizing their inability to produce realistic optical flows. 

\item We demonstrate that image-to-video diffusion models can generate realistic image transformations that preserve contextual relationships, enabling the synthesis of plausible optical flows.

\item We show that these image-flow pairs significantly improve VSOD model performance, achieving state-of-the-art results across public benchmark datasets.
\end{itemize}

\begin{figure*}[t]
\centering
\includegraphics[width=0.88\linewidth]{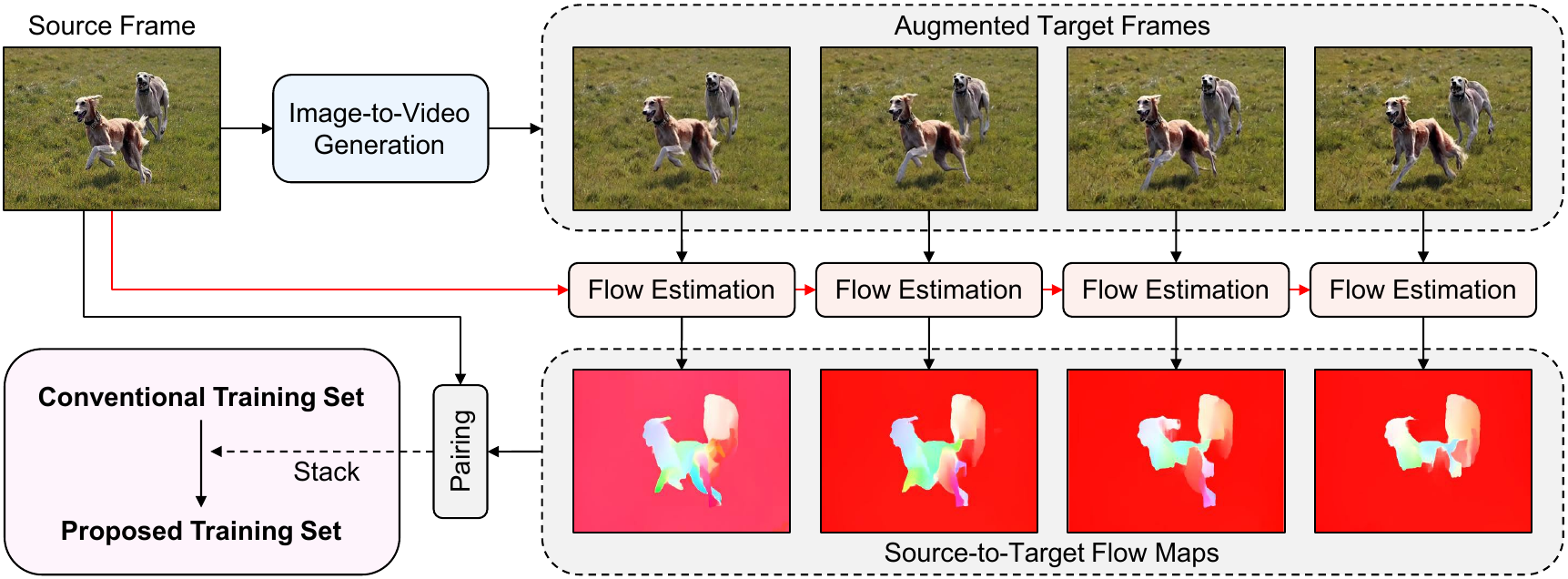}
\caption{Overview of our training sample simulation process. Starting with a static source image, target images are generated using an image-to-video generation model. Optical flow maps are then estimated between the source and target images, with each map paired with the source image to enrich the training dataset.}
\label{figure2}
\end{figure*}

\section{Related Work}

\noindent\textbf{Utilizing image data for video models.}
To leverage the extensive knowledge embedded in large-scale image datasets, video processing tasks commonly incorporate image data. A straightforward approach is to augment image data with spatial transformations and treat the original and newly generated images as a video sequence.

In video object segmentation (VOS), for instance, MaskTrack~\cite{MaskTrack} learns a semi-supervised VOS network by applying affine transformations and thin-plate spline~\cite{TPS} deformations to the segmentation mask of the query frame. The transformed mask is treated as the mask of the previous frame and used along with the query frame image as a training sample. RGMP~\cite{RGMP} adopts a similar approach to MaskTrack, simulating the previous frame’s masks through transformations, while further enriching the diversity of training samples by overlaying foreground objects from salient object detection datasets~\cite{SOD1, SOD2} onto background images from PASCAL-VOC~\cite{PASCAL1, PASCAL2}. To create varied training samples, spatial transformations are separately applied to each image, simulating diverse scenarios. Recent methods leverage image data even more directly. Rather than simply approximating conditions encountered during inference, as MaskTrack and RGVI do, these methods generate video-like samples by independently transforming images and their corresponding object masks, then sequencing them temporally. This approach has been shown to be effective across various VOS settings, including both semi-supervised and unsupervised VOS methods~\cite{STM, KMN, HMMN, STCN, TBD, XMem, 3DC-Seg}.

Beyond VOS, using image data is also common in other video tasks. STEm-Seg~\cite{STEm-Seg} trains a video instance segmentation model using image instance segmentation datasets, synthesizing training clips with random affine transformations and motion blur effects applied on the fly. FCNS~\cite{FCNS} introduces a novel data augmentation technique that simulates video training data from images to learn varied saliency information and mitigate overfitting given the limited availability of training videos.

While these methods effectively implement an image-to-video simulation approach, they rely solely on spatial transformations without accounting for contextual information within each image. As a result, the optical flows generated between the original and transformed images lack meaningful motion cues that could enhance object-level semantics.

\vspace{1mm}
\noindent\textbf{Motion-guided architecture.}
Leveraging motion cues is essential in many video tasks, as it provides crucial insights into the dynamics of each element within images. These cues are typically captured by estimating movement between consecutive frames, incorporating short-term temporal information. This approach is particularly effective in segmentation tasks, as the spatial details preserved in flow maps improve model accuracy in pixel-level prediction.

MATNet~\cite{MATNet} integrates motion information with appearance features through an asymmetric attention block within each encoding layer, transforming appearance features into motion-attentive representations. MGA~\cite{MGA} employs separate modules for salient object detection in still images and motion saliency detection in optical flow maps, adapting these modules to each other through end-to-end training. AMC-Net~\cite{AMC-Net} dynamically balances appearance and motion cues when fusing features from each modality, adjusting feature scaling via gating functions. FSNet~\cite{FSNet} models deep relationships between appearance and motion cues within feature embedding layers, enforcing mutual constraints that enable robust encoding of each modality. PMN~\cite{PMN}, GSA-Net~\cite{GSA-Net}, and DPA~\cite{DPA} leverage attention mechanisms to fuse appearance and motion cues while capturing temporal relationships along the sequence.

Despite these advancements, motion-guided methods often face limitations due to the scarcity of video training data. While image-to-video simulation techniques strive to produce realistic videos, the optical flows generated from this simulated data lack authentic motion characteristics, reducing their effectiveness for training motion-guided models.

\vspace{1mm}
\noindent\textbf{Image-to-video generation.}
Recently, video generation models have gained significant attention due to their wide-ranging applications in content creation and editing. Among these, image-to-video generation models, which produce videos from single images, have demonstrated superior generation quality and controllability compared to general video generation approaches. By using the given image as an explicit first-frame guide, these models ensure high-quality generation aligned with the desired content.

The advent of diffusion models~\cite{DDPM, DDIM, LDM} has greatly advanced image-to-video generation, harnessing the powerful generative capabilities of diffusion. Notable examples of such approaches include Stable Video Diffusion~\cite{SVD}, Gen-2~\cite{Gen-2}, I2VGen-XL~\cite{I2VGen-XL}, PikaLabs~\cite{PikaLabs}, SparseCtrl~\cite{SparseCtrl}, and SORA~\cite{SORA}. Extending these works, some recent methods add more control by explicitly guiding generation with optical flow or trajectory information~\cite{DragNUWA, MoVideo}.

Our objective, however, is purely to transform given images into plausible optical flows. For this purpose, we adopt a straightforward image-to-video architecture that generates natural videos starting from a single image as the first frame, without explicit guidance in the video generation process.

\section{Approach}

\subsection{Transforming Static Images}
As shown in Figure~\ref{figure1}, spatially distorting the original source image to generate a target frame for optical flow estimation does not yield realistic optical flow maps. To obtain high-quality optical flow maps with video-like properties from static images, we use an image-to-video generation model that generates new frames directly, rather than relying on the distortion of the source image.

\vspace{1mm}
\noindent\textbf{Network overview.}
In the image-to-video generation process, we employ Stable Video Diffusion~\cite{SVD} based on a 3D-UNet~\cite{3D-UNet} architecture. Following recent approaches, a pixel-to-latent conversion is applied before the denoising model, with a latent-to-pixel conversion following it, using a pre-trained VAE~\cite{VAE} encoder and decoder, respectively. Notably, the VAE decoder incorporates 3D convolutional layers to account for the time dimension, thereby ensuring temporal consistency during the decoding process.

Starting from Gaussian noise $x \in \mathbb{R}^{C \times THW}$ and the source image latent $z_s \in \mathbb{R}^{C \times HW}$, the denoised latent $z_t \in \mathbb{R}^{C \times THW}$ is obtained as
\begin{align}
&z_t = \Phi(x, \text{repeat}(z_s,T))~,
\end{align}
where $\Phi$ denotes the denoising 3D-UNet model. The output denoised latent $z_t$ is then decoded with the VAE decoder on a per-frame basis, with each frame representing a transformed image generated from the source image. Note that the number of video frames $T$ is set to $14$, following the default setting in Stable Video Diffusion.

\vspace{1mm}
\noindent\textbf{Sampling details.}
During the sampling process, we use $25$ steps of the deterministic DDIM sampler~\cite{detDDIM}. The classifier guidance scale~\cite{CFG} is set to $3.0$ for the first frame and $1.0$ for the last frame. The resolution and frame rate are set to $576 \times 1024$ and $7$, in line with the default setting. The decoding chunk size is set to $8$ to balance generation quality and computational cost.

\subsection{Hallucinating Flows}
After generating target images from each source image using the image-to-video generation model, we obtain sets of source-target pairs for network training. Formally, the temporary paired data $P$ can be represented as
\begin{align}
&P^T_1 = \{(I_s,I_1), (I_s,I_2), ..., (I_s,I_T)\}~,
\end{align}
where $I_s$ represents the source image and $I_t$ denotes the generated target image at each frame $t \in \{1, 2, ..., T\}$. From these temporary pairs, we construct training samples for network training, as illustrated in Figure~\ref{figure2}.

\vspace{1mm}
\noindent\textbf{Data creation.}
In order to obtain the final image-flow pairs to learn the two-stream network for VSOD, we need to estimate flows from the temporary paired data $P$. For optical flow estimation, we use a pre-trained RAFT model~\cite{RAFT} while maintaining the original image resolution. Formally, the final paired data $Q$ can be obtained as 
\begin{align}
&Q^T_1 = \{(I_s,F_{s \rightarrow 1}), (I_s,F_{s \rightarrow 2}), ..., (I_s,F_{s \rightarrow T})\}~,
\end{align}
where $F_{A \rightarrow B}$ indicates the estimated optical flow map from frame $A$ to frame $B$. Note that the paired data $Q$ is obtained for each static image separately, and therefore, from $N$ source images, we can obtain $NT$ paired data samples.

\vspace{1mm}
\noindent\textbf{Dataset construction.}
We adopt the DUTS~\cite{DUTS} dataset as the source for applying our data creation protocol, which consists of 10,553 training images and 5,019 testing images. Since these image samples are used solely for network training, we utilize the entire training and testing sets to construct our training dataset. For each static image from DUTS, $T$ image-flow samples are extracted, resulting in a total of 15,572 data pairs for training. Note that, as the flows are calculated using the source image as the starting frame for optical flow estimation, the RGB images, segmentation masks, and optical flows are all spatially aligned. These data pairs are then combined with conventional training pairs to enhance data diversity during network training.

\begin{figure}[t]
\centering
\includegraphics[width=0.8\linewidth]{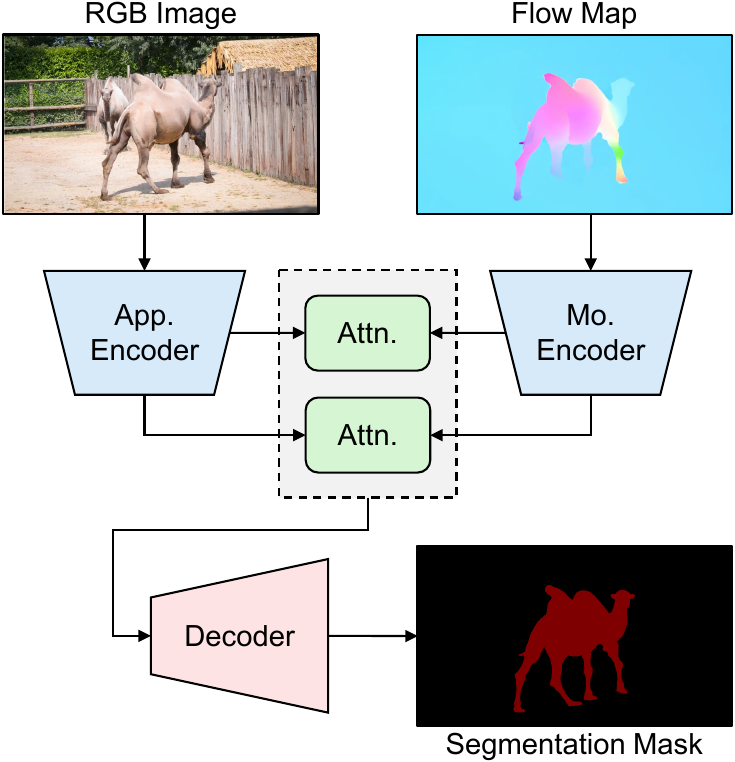}
\caption{Visualization of our two-stream network architecture, which uses both RGB images and optical flow maps as input for primary object mask prediction.}
\label{figure3}
\end{figure}

\subsection{Segmentation Model}
With the aid of our simulated training samples, we train a VSOD network to detect primary objects at the pixel level. Following conventional flow-guided object segmentation approaches, we design our model using a two-stream architecture, as shown in Figure~\ref{figure3}.

\vspace{1mm}
\noindent\textbf{Network architecture.}
Our network includes dual encoders designed to capture appearance cues from RGB images and motion cues from optical flow maps. Both encoders are based on MiT-b2~\cite{MiT}, utilizing only its encoder component. After embedding each modality, multi-level features are fused through simple attention blocks~\cite{CBAM}. The fused features are then passed through the decoder, where they are progressively decoded by combining feature interpolation with skip connections. Following the decoding process, a binary segmentation mask representing the primary objects is obtained.

\vspace{1mm}
\noindent\textbf{Network training.}
The segmentation network is trained using a straightforward supervised learning protocol. We employ cross-entropy loss and the Adam optimizer~\cite{adam}, with a fixed learning rate of 1e-5, a batch size of 16, and an input resolution of $512 \times 512$. The training is conducted on two GeForce RTX TITAN GPUs, each utilizing less than 16GB of GPU memory. The entire network training process takes approximately two days on this hardware. During training, we combine both actual video data and simulated data derived from static images. Specifically, the actual data consists of the DAVIS 2016~\cite{DAVIS} and DAVSOD~\cite{SSAV} training sets. The data samples are randomly shuffled, with the mixture ratio set to 2:1:1 for simulated data, DAVIS 2016 data, and DAVSOD data, respectively.

\begin{table*}[t]
\centering 
\caption{Quantitative evaluation on the DAVIS 2016 validation set, FBMS test set, DAVSOD test set, and ViSal dataset. Scores marked in \textcolor{red}{red} indicate the highest score, while scores marked in \textcolor{blue}{blue} represent the second-highest score.}
\vspace{-2mm}
\small
\begin{tabular}{p{2.0cm}|P{5.8mm}P{5.8mm}P{5.8mm}|P{5.8mm}P{5.8mm}P{5.8mm}|P{5.8mm}P{5.8mm}P{5.8mm}|P{5.8mm}P{5.8mm}P{5.8mm}|P{5.8mm}P{5.8mm}P{5.8mm}}
\toprule
\multicolumn{1}{c}{} &\multicolumn{3}{c}{\cellcolor{Gray}DAVIS 2016} &\multicolumn{3}{c}{FBMS} &\multicolumn{3}{c}{\cellcolor{Gray}DAVSOD} &\multicolumn{3}{c}{ViSal} &\multicolumn{3}{c}{\cellcolor{Gray}Average}\\
Method &$\mathcal{S}\uparrow$ &$\mathcal{F}\uparrow$ &$\mathcal{M}\downarrow$ &$\mathcal{S}\uparrow$ &$\mathcal{F}\uparrow$ &$\mathcal{M}\downarrow$ &$\mathcal{S}\uparrow$ &$\mathcal{F}\uparrow$ &$\mathcal{M}\downarrow$ &$\mathcal{S}\uparrow$ &$\mathcal{F}\uparrow$ &$\mathcal{M}\downarrow$  &$\mathcal{S}\uparrow$ &$\mathcal{F}\uparrow$ &$\mathcal{M}\downarrow$\\
\midrule
SSAV~\cite{SSAV} &89.3 &86.1 &2.8 &87.9 &86.5 &4.0 &72.4 &60.3 &9.2 &94.3 &93.9 &2.0 &86.0 &81.7 &4.5\\
AD-Net~\cite{AD-Net} &- &80.8 &4.4 &- &81.2 &6.4 &- &- &- &- &90.4 &3.0 &- &- &-\\
F$^3$Net~\cite{F3Net} &85.0 &81.9 &4.1 &85.3 &81.9 &6.8 &68.9 &56.4  &11.7 &87.4 &90.7 &4.5 &81.7 &77.7 &6.8\\
MINet~\cite{MINet} &86.1 &83.5 &3.9 &84.9 &81.7 &6.7 &70.4 &58.2 &10.3 &90.3 &91.1 &4.1 &82.9 &78.6 &6.3\\
GateNet~\cite{GateNet} &86.9 &84.6 &3.6 &85.7 &83.2 &6.5 &70.1 &57.8 &10.4 &92.1 &92.8 &3.9 &83.7 &79.6 &6.1\\
PCSA~\cite{PCSA} &90.2 &88.0 &2.2 &86.6 &83.1 &4.1 &74.1 &65.5 &8.6 &94.6 &94.0 &1.7 &86.4 &82.7 &4.2\\
DFNet~\cite{DFNet} &- &89.9 &1.8 &- &83.3 &5.4 &- &- &- &- &92.7 &1.7 &- &- &-\\
3DC-Seg~\cite{3DC-Seg} &- &91.8 &1.5 &- &84.5 &4.8 &- &- &- &- &92.2 &1.9 &- &- &-\\
CASNet~\cite{CASNet} &87.3 &86.0 &3.2 &85.6 &86.3 &5.6 &69.4 &- &8.9 &82.0 &84.7 &2.9 &81.1 &- &5.2\\
FSNet~\cite{FSNet} &92.0 &90.7 &2.0 &89.0 &88.8 &4.1 &77.3 &68.5 &7.2 &- &- &- &- &- &-\\
CFAM~\cite{CFAM} &91.8 &90.9 &1.5 &90.9 &\textcolor{blue}{91.5} &\textcolor{blue}{2.6} &75.3 &66.2 &8.3 &94.7 &95.1 &1.3 &88.2 &85.9 &3.4\\
UFO~\cite{UFO} &91.8 &90.6 &1.5 &89.1 &88.8 &3.1 &- &- &- &\textcolor{blue}{95.9} &95.1 &1.3 &- &- &-\\
DBSNet~\cite{DBSNet} &92.4 &91.4 &1.4 &88.2 &88.5 &3.8 &77.8 &68.8 &7.6 &93.1 &92.8 &2.0 &87.9 &85.4 &3.7\\
HFAN~\cite{HFAN} &93.4 &92.9 &\textcolor{red}{0.9} &87.5 &84.9 &3.3 &75.3 &68.0 &7.0 &94.1 &93.5 &\textcolor{blue}{1.1} &87.6 &84.8 &3.1\\
TMO~\cite{TMO} &92.8 &92.0 &\textcolor{red}{0.9} &88.6 &88.2 &3.1 &76.7 &70.8 &7.2 &94.2 &94.7 &\textcolor{red}{1.0} &88.1 &86.4 &3.1\\
OAST~\cite{OAST} &\textcolor{blue}{93.5} &\textcolor{blue}{92.6} &1.1 &\textcolor{blue}{91.7} &\textcolor{red}{91.9} &\textcolor{red}{2.5} &78.6 &71.2 &7.0 &94.8 &95.0 &\textcolor{red}{1.0} &89.7 &87.7 &\textcolor{blue}{2.9}\\
TGFormer~\cite{TGFormer} &93.2 &92.2 &1.1 &91.6 &\textcolor{red}{91.9} &\textcolor{blue}{2.6} &\textcolor{blue}{79.8} &\textcolor{blue}{72.8} &\textcolor{red}{6.5} &95.2 &\textcolor{blue}{95.5} &\textcolor{blue}{1.1} &\textcolor{blue}{90.0} &\textcolor{blue}{88.1} &\textcolor{red}{2.8}\\
\midrule
\textbf{RealFlow} &\textcolor{red}{94.5} &\textcolor{red}{93.9} &\textcolor{blue}{1.0} &\textcolor{red}{92.6} &90.6 &2.8 &\textcolor{red}{80.3} &\textcolor{red}{73.2} &\textcolor{blue}{6.6} &\textcolor{red}{96.2} &\textcolor{red}{96.6} &\textcolor{red}{1.0} &\textcolor{red}{90.9} &\textcolor{red}{88.6} &\textcolor{blue}{2.9}\\
\bottomrule
\end{tabular}
\label{table1}
\end{table*}

\section{Experiments}
This section outlines the datasets and evaluation metrics used to validate our approach. We then quantitatively and qualitatively compare our method to other state-of-the-art methods and analyze it through additional experiments. All evaluations are conducted on a single GeForce RTX 2080 Ti GPU. Our VSOD network, trained with realistic simulated image-flow pairs, is referred to as RealFlow.

\subsection{Datasets}
We use four widely adopted VSOD datasets: DAVIS 2016~\cite{DAVIS}, FBMS~\cite{FBMS}, DAVSOD~\cite{SSAV}, and ViSal~\cite{ViSal}. For each video, optical flows are precomputed between consecutive frames using RAFT~\cite{RAFT} and saved alongside the RGB images and segmentation masks.

\vspace{1mm}
\noindent\textbf{DAVIS 2016.} A primary VSOD benchmark, the DAVIS dataset includes 50 videos with densely annotated segmentation masks for every frame. Of these, 30 videos are designated for training and 20 for validation. We incorporate the training set into our data for model training.

\vspace{1mm}
\noindent\textbf{FBMS.} The FBMS dataset contains 59 videos, with 29 for training and 30 for validation. Due to sparse frame annotations, only the annotated frames are used. Consistent with VSOD protocols, we exclude its training set from our data.

\vspace{1mm}
\noindent\textbf{DAVSOD.} The DAVSOD dataset, a recent benchmark in VSOD, comprises 61 training videos and 35 test videos, with full annotations for each frame. We include its training set in our training data.

\vspace{1mm}
\noindent\textbf{ViSal.} The ViSal dataset consists of 17 videos with 193 annotated frames, though only selected frames are labeled. Since it lacks a defined training and testing split, we use it solely for evaluation.

\begin{figure*}[t]
\centering
\includegraphics[width=1.0\linewidth]{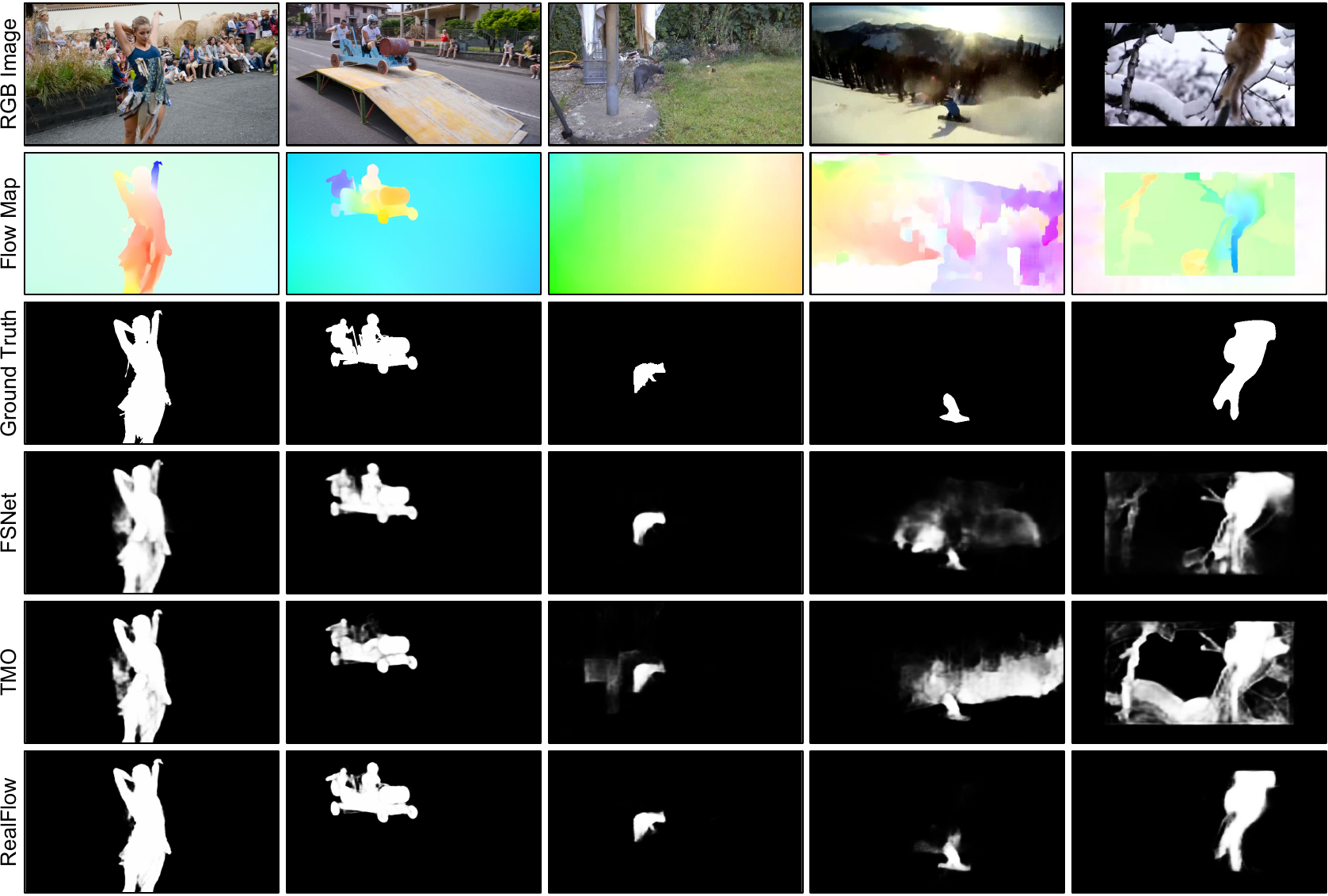}
\caption{Qualitative comparison between state-of-the-art methods and the proposed RealFlow.}
\label{figure4}
\end{figure*}

\subsection{Evaluation Metrics}
We adopt three evaluation metrics to quantitatively validate the effectiveness of our approach: $\mathcal{S}$~\cite{smeasure}, $\mathcal{F}$, and $\mathcal{M}$.

\vspace{1mm}
\noindent\textbf{$\mathcal{S}$ measure.} The $\mathcal{S}$ measure evaluates structural similarity between the predicted saliency map and the ground truth segmentation mask, calculated as 
\begin{align} 
&\mathcal{S} = \alpha \mathcal{S}_o + (1 - \alpha) \mathcal{S}_r~, 
\end{align} 
where $\mathcal{S}_o$ represents object-aware structural similarity and $\mathcal{S}_r$ represents region-aware structural similarity. Following existing approaches, $\alpha$ is set to 0.5.

\vspace{1mm}
\noindent\textbf{$\mathcal{F}$ measure.} The $\mathcal{F}$ measure is based on the precision and recall of the predicted saliency map, calculated as 
\begin{align} 
&\mathcal{F} = \frac{(1 + \beta^2) \times \text{Precision} \times \text{Recall}}{\beta^2 \times \text{Precision} + \text{Recall}}~, 
\end{align} 
where $\beta^2$ is set to 0.3 following \cite{fmeasure}.

\vspace{1mm}
\noindent\textbf{$\mathcal{M}$ measure.} The $\mathcal{M}$ measure calculates the average pixel-wise difference between the saliency prediction and the ground truth mask. Formally, it is represented as 
\begin{align} 
&\mathcal{M} = \frac{\textstyle\sum_{p=1}^H \textstyle\sum_{q=1}^W{|M_{pred}^{p,q} - M_{gt}^{p,q}|}}{H \times W}~, 
\end{align} 
where $M_{pred}$ and $M_{gt}$ denote the predicted saliency map and the ground truth mask, respectively; $H$ and $W$ represent the image height and width, and $p$ and $q$ indicate individual pixel locations along the height and width dimensions.

\subsection{Quantitative Results}
We compare our approach against existing state-of-the-art methods on the DAVIS 2016~\cite{DAVIS} validation set, FBMS~\cite{FBMS} test set, DAVSOD~\cite{SSAV} test set, and ViSal~\cite{ViSal} dataset, as presented in Table~\ref{table1}. For a comprehensive view, we also report the average score across these four datasets for each metric. Our method achieves the highest $\mathcal{S}$ measure~\cite{smeasure} score on all benchmark datasets, surpassing prior methods by a substantial margin. The $\mathcal{F}$ measure performance is similarly strong, achieving the top score on three of the four benchmarks. For the $\mathcal{M}$ measure, our approach demonstrates competitive performance, reaching accuracy levels comparable to other leading methods. These results indicate that our method offers significant improvements over existing VSOD solutions, delivering high saliency prediction accuracy across all metrics and establishing its effectiveness across diverse scenarios.

\subsection{Qualitative Results}
In Figure~\ref{figure4}, we compare our approach with state-of-the-art methods FSNet~\cite{FSNet} and TMO~\cite{TMO} across varied VSOD scenarios. In the first two columns, where primary objects are well-aligned with optical flow maps, our method delivers sharper and more precise segmentation, especially along object edges, surpassing existing methods. The third column shows cases with minimal object movement relative to the background, where our network effectively leverages RGB appearance cues to accurately capture primary objects. The fourth and fifth examples depict scenes with substantial motion and noisy optical flow maps, which confuse other methods. In contrast, our approach remains robust, accurately segmenting primary target objects even in complex motion scenarios.

\begin{figure*}[t]
\centering
\includegraphics[width=1.0\linewidth]{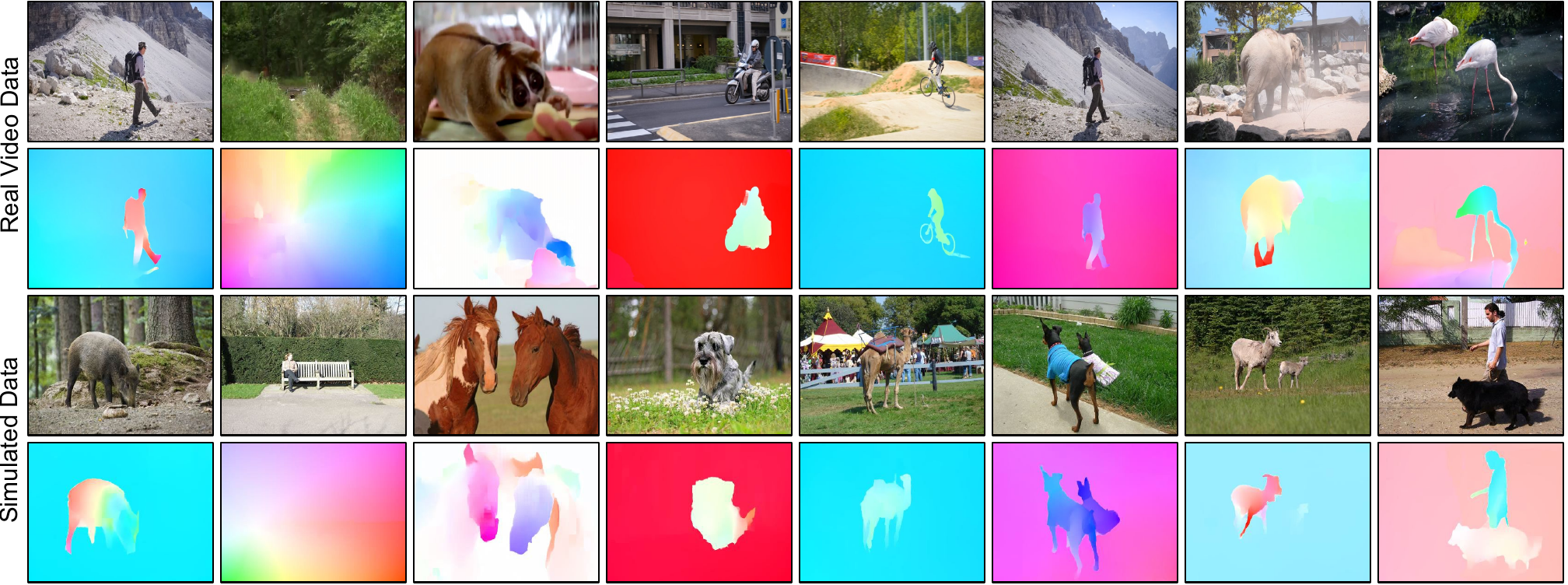}
\caption{Aligned qualitative comparison between real video data and our simulated data.}
\label{figure5}
\end{figure*}

\begin{figure}[t]
\centering
\includegraphics[width=1.0\linewidth]{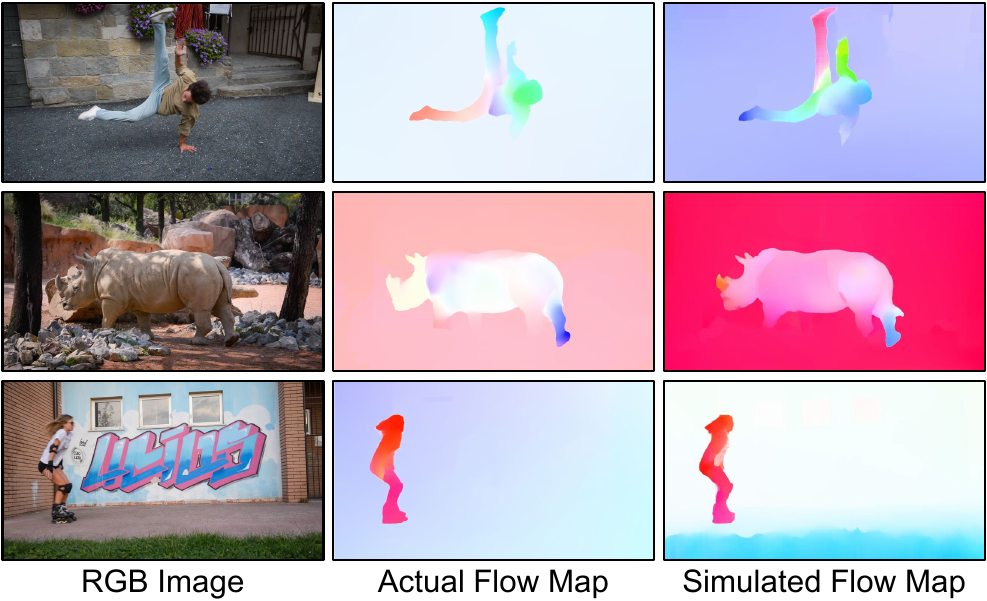}
\caption{Direct qualitative comparison of actual and simulated flow maps, both derived from the same source image.}
\label{figure6}
\end{figure}

\subsection{Analysis}
In this section, we conduct a series of experiments to provide a deeper analysis of the proposed approach, offering additional insights and intuitions into its effectiveness. These experiments aim to deliver a comprehensive understanding of the key properties of our method, highlighting its strengths, limitations, and the underlying mechanisms driving its performance.

\vspace{1mm}
\noindent\textbf{Comparison to real data.}
The goal of our approach is to simulate realistic image-flow pairs that closely mirror those from real video data, thereby improving network training. To evaluate the plausibility of the simulated data, we provide an aligned comparison with real video data in Figure~\ref{figure5}. As seen in various examples, the simulated data contains a range of optical flow maps, from explicit flows that capture object shapes to more ambiguous flows with less distinct object information. This demonstrates that the simulated data spans a wide distribution of scenarios, supporting the validity of incorporating it into network training.

Furthermore, we qualitatively compare the actual flow map with our simulated flow map, using the same source image, in Figure~\ref{figure6}. The comparison shows that the simulated flow map effectively replicates the desired object movements within the image, generating plausible optical flow maps for VSOD. This further substantiates the effectiveness and validity of our data simulation strategy.

\vspace{1mm}
\noindent\textbf{Training protocol.}
We validate the effectiveness of our simulated data for network training in Table~\ref{table2}. Notably, the network trained solely on our simulated data outperforms the network trained on real video data, such as the DAVIS 2016~\cite{DAVIS} and DAVSOD~\cite{SSAV} training sets, on all datasets except the DAVIS 2016 validation set. This result highlights the limitations of existing training data protocols and underscores the quality and effectiveness of our simulated data in building a robust VSOD model. When we combine real video data with our simulated data, we achieve the highest performance across all datasets and metrics, emphasizing the critical role of training data volume in network training.

\begin{table*}[t]
\centering 
\caption{Ablation study on the training protocol.}
\vspace{-2mm}
\small
\begin{tabular}{P{1.7cm}|P{5.8mm}P{5.8mm}P{5.8mm}|P{5.8mm}P{5.8mm}P{5.8mm}|P{5.8mm}P{5.8mm}P{5.8mm}|P{5.8mm}P{5.8mm}P{5.8mm}|P{5.8mm}P{5.8mm}P{5.8mm}}
\toprule
\multicolumn{1}{c}{} &\multicolumn{3}{c}{\cellcolor{Gray}DAVIS 2016} &\multicolumn{3}{c}{FBMS} &\multicolumn{3}{c}{\cellcolor{Gray}DAVSOD} &\multicolumn{3}{c}{ViSal} &\multicolumn{3}{c}{\cellcolor{Gray}Average}\\
Training &$\mathcal{S}\uparrow$ &$\mathcal{F}\uparrow$ &$\mathcal{M}\downarrow$ &$\mathcal{S}\uparrow$ &$\mathcal{F}\uparrow$ &$\mathcal{M}\downarrow$ &$\mathcal{S}\uparrow$ &$\mathcal{F}\uparrow$ &$\mathcal{M}\downarrow$ &$\mathcal{S}\uparrow$ &$\mathcal{F}\uparrow$ &$\mathcal{M}\downarrow$ &$\mathcal{S}\uparrow$ &$\mathcal{F}\uparrow$ &$\mathcal{M}\downarrow$\\
\midrule
Real &93.1 &91.6 &1.3 &85.8 &85.5 &5.3 &76.0 &68.7 &7.8 &94.5 &94.4 &1.8 &87.4 &85.1 &4.1\\
Simulated &91.7 &89.6 &2.0 &91.0 &89.0 &2.9 &78.1 &70.1 &7.8 &96.1 &96.4 &1.1 &89.2 &86.3 &3.5\\
Mixed &94.5 &93.9 &1.0 &92.6 &90.6 &2.8 &80.3 &73.2 &6.6 &96.2 &96.6 &1.0 &90.9 &88.6 &2.6\\
\bottomrule
\end{tabular}
\label{table2}
\end{table*}

\begin{table}[t]
\centering 
\caption{Ablation study on the backbone network.}
\vspace{-2mm}
\small
\begin{tabular}{P{1.4cm}|P{0.6cm}P{1.0cm}P{0.6cm}P{0.6cm}P{0.6cm}}
\toprule
Backbone &fps &Param~\# &$\mathcal{S}\uparrow$ &$\mathcal{F}\uparrow$ &$\mathcal{M}\downarrow$\\
\midrule
MiT-b0 &61.1 &13.8M &89.4 &87.0 &3.5\\
MiT-b1 &45.5 &33.6M &89.8 &87.3 &3.2\\
MiT-b2 &29.5 &55.7M &90.9 &88.6 &2.6\\
\bottomrule
\end{tabular}
\label{table3}
\end{table}

\vspace{1mm}
\noindent\textbf{Backbone network.}
Table~\ref{table3} presents a quantitative comparison of different backbone network versions, with evaluation scores averaged across four datasets: DAVIS 2016~\cite{DAVIS} validation set, FBMS~\cite{FBMS} test set, DAVSOD~\cite{SSAV} test set, and ViSal~\cite{ViSal} dataset. Our default backbone, MiT-b2~\cite{MiT}, achieves the highest scores across all evaluation metrics, striking an effective balance of inference speed and parameter efficiency suitable for real-time applications. Lighter backbones provide even faster inference speeds and reduced parameter counts, with a minor accuracy trade-off, offering flexibility for applications that prioritize speed or precision. The consistent trend of better scores with larger backbones further reinforces the stability of our method.

\vspace{1mm}
\noindent\textbf{Input resolution.}
Table~\ref{table4} compares various input resolution settings to evaluate our method in different scenarios. Similar to Table~\ref{table3}, average metric scores are calculated across the four datasets. Each network is re-trained and tested at its specific resolution setting to ensure accurate results. As shown, lower resolutions yield faster inference speeds with a slight decrease in prediction accuracy. Like the backbone choice, input resolution can be adjusted to balance speed and accuracy based on user needs, making the method adaptable to various application requirements.

\begin{table}[t]
\centering 
\caption{Ablation study on the input resolution.}
\vspace{-2mm}
\small
\begin{tabular}{P{1.7cm}|P{0.6cm}P{0.6cm}P{0.6cm}P{0.6cm}}
\toprule
Resolution &fps &$\mathcal{S}\uparrow$ &$\mathcal{F}\uparrow$ &$\mathcal{M}\downarrow$\\
\midrule
256$\times$256 &55.0 &90.0 &87.5 &3.0\\
384$\times$384 &44.2 &90.2 &87.8 &2.9\\
512$\times$512 &29.5 &90.9 &88.6 &2.6\\
\bottomrule
\end{tabular}
\label{table4}
\end{table}

\begin{figure}[t]
\centering
\includegraphics[width=1.0\linewidth]{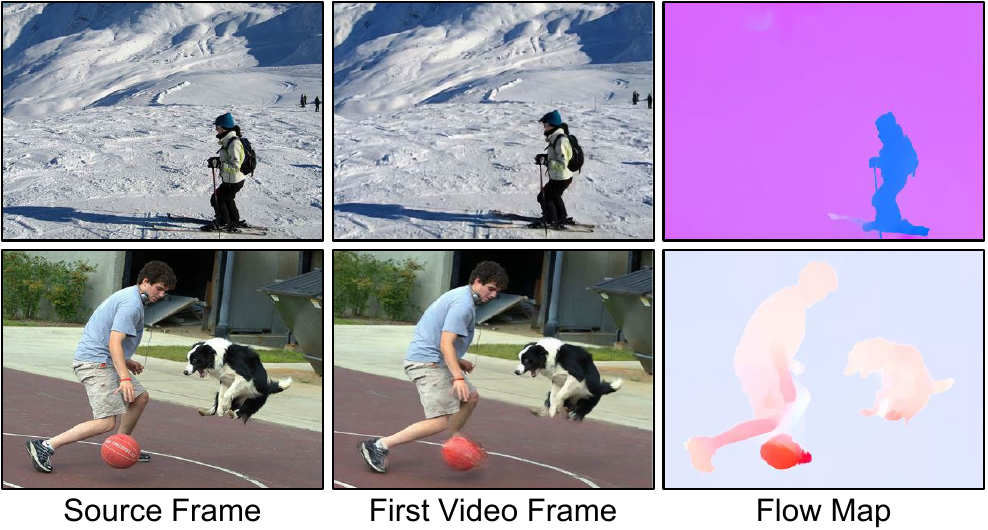}
\caption{Visualization of the flow map generated between the source image and the first frame of the video.}
\label{figure7}
\end{figure}

\vspace{1mm}
\noindent\textbf{Source image deformation.}
The image-to-video generation model is trained to produce a plausible video sequence, with the given source image serving as the first frame. While the source image implicitly guides the generation process, this constraint is not strictly enforced, so the generated first frame may slightly differ from the source image in some instances, as shown in Figure~\ref{figure7}. Although this variation could indicate suboptimal generation quality, it is not considered a failure in the context of flow simulation. Instead, the generated first frame can still be paired with the source image for optical flow estimation, contributing to the diversity of the training data.

\begin{figure}[t]
\centering
\includegraphics[width=1.0\linewidth]{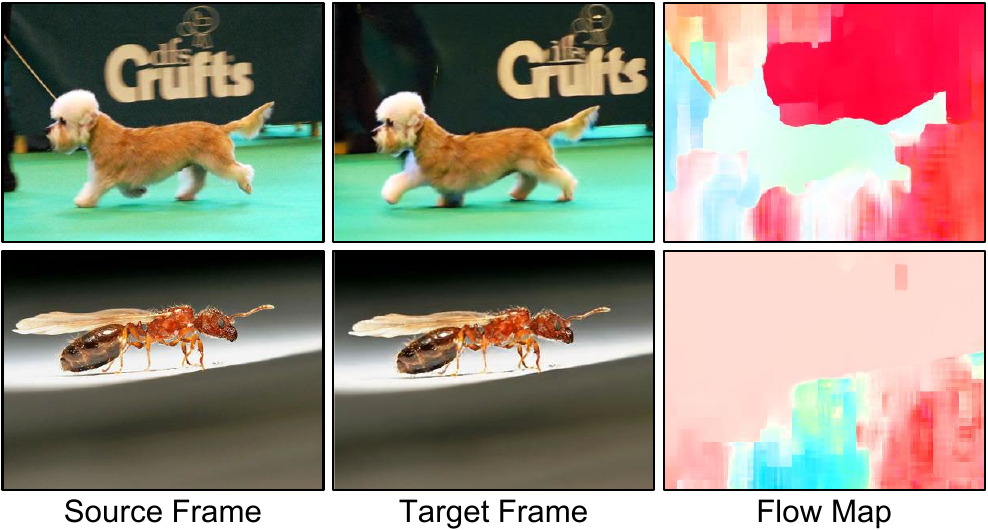}
\caption{Visualization of the limitations of our simulated flows.}
\label{figure8}
\end{figure}

\vspace{1mm}
\noindent\textbf{Limitation.}
The image-to-video generation process does not always fully replicate real-world video dynamics. For example, the generated video frames often lack continuous motion, resulting in unnatural motion dynamics. As shown in Figure~\ref{figure8}, optical flow maps frequently exhibit checkerboard artifacts. This issue may stem from the patch-level copying of content between frames within the video generation model, possibly due to limitations in the video simulation process or the latent decoding mechanism.

Moreover, while multiple video frames are generated, they are used solely as target frames for optical flow generation, with the original image serving as the source frame. This limitation arises because only the transformed RGB images are available, preventing the newly generated frames from being used as the source for optical flow calculation. If transformed segmentation masks could be generated concurrently with the RGB images during the image-to-video generation process, it would enable the creation of more diverse training data.

\section{Conclusion}
In this study, we investigate the use of large-scale image data to train flow-guided models. Rather than relying on source image distortion methods that fail to accurately mimic the optical flows of real video data, we demonstrate that novel image generation through image-to-video generation provides an effective solution. For the VSOD task, our approach sets a new state-of-the-art performance across all benchmark datasets, surpassing existing methods.

{
    \small
    \bibliographystyle{ieeenat_fullname}
    \bibliography{RealFlow}
}

\end{document}